\documentclass[final,3p,times,twocolumn]{elsarticle}

\usepackage{graphicx}
\usepackage{subfigure}
\usepackage{booktabs}
\usepackage{multirow}
\usepackage{amssymb}
\usepackage{amsmath,amssymb}

\journal{Signal Processing: Image Communication}

\begin{document}

\begin{frontmatter}



\title{SFA: Small Faces Attention Face Detector}

\address[add1]{Key Laboratory of Symbol Computation and Knowledge Engineer of Ministry of Education, Jilin University, Changchun 130012, China}
\address[add2]{College of Computer Science and Technology, Jilin University, Changchun 130012, China}

\author[add1,add2]{Shi Luo}\ead{shiluo1990@hotmail.com}

\author[add1,add2]{Xiongfei Li}\ead{lxf@jlu.edu.cn}

\author[add1,add2]{Xiaoli Zhang\corref{cor1}}\ead{xiaolizhang1987@gmail.com}

\cortext[cor1]{Corresponding author at: College of Computer Science and Technology, Jilin University, Changchun, China}

\begin{abstract}
In recent year, tremendous strides have been made in face detection thanks to deep learning. However, most published face detectors deteriorate dramatically as the faces become smaller. In this paper, we present the Small Faces Attention (SFA) face detector to better detect faces with small scale. First, we propose a new scale-invariant face detection architecture which pays more attention to small faces, including 4-branch detection architecture and small faces sensitive anchor design. Second, feature maps fusion strategy is applied in SFA by partially combining high-level features into low-level features to further improve the ability of finding hard faces. Third, we use multi-scale training and testing strategy to enhance face detection performance in practice. Comprehensive experiments show that SFA significantly improves face detection performance, especially on small faces. Our real-time SFA face detector can run at 5 FPS on a single GPU as well as maintain high performance. Besides, our final SFA face detector achieves state-of-the-art detection performance on challenging face detection benchmarks, including WIDER FACE and FDDB datasets, with competitive runtime speed. Both our code and models will be available to the research community.
\end{abstract}


\end{frontmatter}


\section{Introduction}
Face detection is a fundamental step of many face related applications, such as face alignment [1, 2], face recognition [3, 4], face verification [5, 6] and face expression analysis. Excellent face detectors can exactly find and locate faces from an image. In recent years, deep learning methods, especially convolutional neural network (CNN) has achieved remarkable successes in a variety of computer vision tasks, ranging from image classification [7, 8] to object detection [9, 10, 11, 12], which also inspires face detection. Unlike traditional methods of hand-crafted features, CNN based method can extract face features automatically. Anchor-based face detectors play a dominant role in CNN-based face detectors. They detect faces by classifying and regressing a series of pre-set anchors, which are generated by regularly tiling a collection of boxes with different scales and aspect ratios on the images. 
\par Small faces are difficult to be detected due to its small scale. Faces with high detection difficulty are categorized as hard faces. Most of small faces belong to hard faces. However, small scale is just one of those variations making faces hard to be detected. Better tackling hard faces is helpful for detecting small faces.
\par Despite significant progress, there are still relevant open questions in face detection. Specifically, the performance of anchor-based face detectors drops dramatically as the faces become smaller. The size of smallest anchors in previous face detection methods is set to 16 which is still too large to match small faces. To solve this problem, some improvements are applied in our method to better detect small faces. That is our initial motivation. 
\par In this paper, we propose the Small Faces Attention (SFA) face detector to finding more faces with small scale. We first propose 4-branch face detection architecture to deal with large, medium and small faces respectively. In particular, two branches in SFA focus on small faces. Then, we redesign the anchors, named small faces sensitive anchor design, by adding more anchors to match small faces. Besides, feature maps fusion strategy is applied in SFA by partially combining high-level features into low-level features to further improve the ability of detecting hard faces. Hence, only two branches for small faces mentioned above use feature map fusion strategy. Finally, we adopt multi-scale training and testing strategy to enhance the performance of face detection. Though previous face detectors are scale-invariant by design, image pyramid can also improve the performance in both training and testing phase.
\par SFA performs face detection in a single stage by scanning the entire image with a sliding window fashion. It detects faces directly from the early feature maps by classifying a set of predefined anchors and regressing them at the same time. More importantly, SFA can find faces from images with arbitrary size and the runtime of SFA is independent of the number of faces in an image. This is in contrast to proposal-based two stage detectors such as Faster R-CNN [12], which scale linearly with the number of proposals. Meanwhile, SFA is scale-invariant by design. We simultaneously detect faces with multiple scales from different layers in a single forward pass of the network.
\par For clarity, the main contributions of this paper can be summarized as:
\begin{itemize}
	\item We propose a new scale-invariant face detection architecture which pays more attention to small faces, including 4-branch detection architecture and small faces sensitive anchor design.
	\item Feature maps fusion strategy is applied in SFA by partially combining high-level features into low-level features to further improve the ability of detecting hard faces.
	\item We use multi-scale training and testing strategy to enhance face detection performance.
	\item Our method achieves state-of-the-art detection performance on challenging face detection benchmarks, including WIDER FACE and FDDB datasets, with competitive runtime speed.
\end{itemize}
\par The rest of the paper is organized as follows. Section 2 briefly reviews the related work in face detection. Section 3 presents the proposed SFA face detector. Section 4 shows our experimental results. Section 5 concludes this paper.
\section{Related Work}
Face detection is a critical and fundamental step to all facial analysis applications, and has been extensively studied over the past few decades. The existing algorithms can be roughly divided into two categories as follows.
\par Traditional approaches: The milestone work of Viola-Jones [13] used Haar-like feature and AdaBoost to train a cascade of face detector that achieved a good accuracy. After that, many approaches have been proposed based on the Viola-Jones detectors to advance the state-of-the-art in face detection. LBP [14] and its extension methods introduced local texture features for face detection. These features have been proved to be robust to illumination variations. NPDFace [15] was to address challenges in unconstrained face detection, such as arbitrary pose variations and occlusions. All of these detectors extract hand-crafted features and optimize each component separately, which makes these traditional face detectors less optimal.
\par CNN-based approaches: In contrast to traditional face detection approaches, CNN-based face detectors greatly improve the detecting performance in recent years. These methods can train on huge and challenging face datasets and automatically extract discriminative features. Furthermore, they can be easily parallelized on GPU cores for acceleration in testing phase. CascadeCNN [16] developed a cascade architecture built on CNNs to detect face coarse to fine. Faceness [17] trained a series of CNNs for facial attribute recognition to detect partially occluded faces. MTCNN [18] proposed to jointly solve face detection and alignment using several multi-task CNNs. FaceHunter [19] proposed a new multi-task CNNs based face detector to discriminate face/non-face and regress face box.
\par Anchor was first proposed by Faster R-CNN [12], and then it was widely used in both two stage and single stage object detectors. Later, anchor-based detecting methods were applied in face detection leading to a remarkable progress. SSH [20] introduced a single stage headless face detector and modelled the context information by large filters on each prediction module. S$^{3}$FD [21] presented a scale-equitable framework to handle different scales of faces. FaceBoxes [22] introduced anchor densification to ensure different types of anchors have the same density on the image. Face R-CNN [23] employed a new multi-task loss function based on Faster R-CNN framework. CMS-RCNN [24] exploited contextual information to enhance performance. Face R-FCN [25] re-weighted embedding responses on score maps and eliminated the effect of non-uniformed contribution in each facial part.

\par Despite its great achievement, the main drawback of these frameworks is their poor detection performance for faces with small scale. To address this problem, great efforts have been done in this aspect. HR [26] built multi-level image pyramids to find upscaled small faces. S$^{3}$FD [21] proposed anchor matching strategy to improve the recall rate of small faces. Zhu et al. [27] introduced a novel anchor design to guarantee high overlaps between small faces and anchor boxes.
\par Although many face detectors are developed, the detection accuracy is still not satisfied, especially for small faces. In this paper, we are interested in developing efficient face detector to better deal with small faces. To this end, SFA face detector is proposed extending from SSH which is an elegant and efficient detection architecture.
\section{Proposed Method}
\subsection{General Architecture}
The pipeline of face detection using SFA is illustrated in Fig. 1(a). The input image \textit{I} with arbitrary size is resized to form an image collection \textit{P = \{  P$_{1}$, P$_{2}$, $\cdots$, P$_{i}$, $\cdots$,P$_{n}$ \}} according to scale \textit{S = \{ S$_{1}$, S$_{2}$, $\cdots$, S$_{i}$, $\cdots$, S$_{n}$ \}} in Multi-scale Testing. Each image \textit{P$_{i}$} uses SFA to generate detection result \textit{D$_{i}$}. We merge these detection results to get image \textit{D$_{f}$} as our final detection result of input image \textit{I}.
\par Fig. 1(b) shows the network architecture of SFA. First of all, VGG-16 [7] is deployed to extract feature maps from resized image \textit{P$_{i}$}. Then, Feature Maps Fusion strategy is applied to fuse feature maps from Conv3$\_$3, Conv4$\_$3, and Conv5$\_$3. Finally, we use a set of Scale-invariant detection modules to classify face/non-face and regress the bounding boxes. Detection module M0, M1, M2, and M3 detect faces with small, medium, and large scale respectively. We exploit NMS to generate detection result \textit{D$_{i}$} of image \textit{P$_{i}$}.

\subsection{4-branch detection architecture}
Face detection in presence of small faces is an important issue in unconstrained scenarios. However, the intrinsic properties of small faces always make them hard to be detected efficiently. Small faces are small in scale which usually bellows 40 pixels. Meanwhile, anchor-based face detectors based on CNNs exploit multi-layer convolution and pooling operation to extract discriminative face features. Hence, the size of receptive fields become larger gradually as the feature maps are extracted from low-level to high-level as listed in Tab. 1. Thus, the large size of receptive fields challenges the scale of small faces. Low-level feature information of faces is lost gradually when CNNs based feature extraction method is applied. In the end, minority of feature information is preserved for the small faces, which leads to poor performance in detecting small faces. Therefore, it is necessary to detect small faces from early detection layers where still maintain more low-level features.
\par To this end, we propose a new scale-invariant face detection architecture, named 4-branch detection architecture as shown in Fig. 1(b). Inspired by FPN [28], we detect faces from four different layers of VGG-16 using detection modules M0, M1, M2, and M3. Conv3\_3, Conv4\_3, Conv5\_3, and Pool5 are selected to connect to the detection modules M0, M1, M2, and M3 separately. These modules have strides of 4, 8, 16, and 32. And they are designed to detect small, medium, and large faces respectively. In particular, two branches of M0 and M1 in SFA focus on faces with small scale. The detection module of SSH is deployed in our method, which is elegant and efficient. It consists of a convolutional binary classifier and a regressor for detecting faces and localizing them.
\par During the training phase, each detection module \textit{M$_{i}$}, where \textit{i} $\in$ \{0, 1, 2, 3\}, is trained to detect faces from a target scale range. SFA uses a multi-task loss function in the training phase. It is similar to the common formulation but consists of 4 pairs of softmax loss and smooth L1 loss consistent with 4-branch face detection architecture. To specialize each of the four detection modules for a specific range of scales, we only back-propagate the loss for the anchors which are assigned to faces in the corresponding range. This is implemented by distributing the anchors based on their size to these four modules as discussed in Section 3.3. Unlike S$^{3}$FD, which merges different scales of feature maps and forms a comprehensive face features, our work indicates that multi-branch detection modules in scale can be optimally learned separately. In this way, different scales of faces can be automatically divided into different detection modules. This is the divide and conquer strategy to tackle unconstrained face detection in a single detector.
\par During inference, the predicted boxes from the different branches are joined together followed by Non-Maximum Suppression (NMS) to form the final detections.

\subsection{Small Faces Sensitive Anchor Design}
Anchor-based face detection methods can be regards as a binary classification problem, which determines if an anchor is face or not. However, few anchors in previous face detectors are offered to match small faces. For example, the size of smallest anchors in SSH [20], S$^{3}$FD [21] and Zhu et al. [27] is 16. 
\par To better detect small faces, we propose small faces sensitive (SFS) anchor design. We tile anchors on a wide range of size varying from 4 to 512 (i.e., 4, 8, 16, 32, 64, 128, 256, 512 in our method), which guarantees that various scales of faces have enough features for detection. More precisely, as listed in Tab. 2, the smallest anchor in our method is 4. And the anchors of 4, 8, 16, and 32 are applied for faces with small scale. Benefit from the 4-branch detection architecture as discussed in Section 3.2, SFA reasonably arranges small faces sensitive anchors for 4-branch detection modules and forms our SFS anchor design, which improves the robustness to face scales.
\par For implementation, we use anchor ratio (AR) and base size (BS) to form anchor design. AR multiple BS is the size of anchor. The AR of \{1, 2\} in M0, \{4, 8\} in M1, \{16, 32\} in M2, and \{64, 128\} in M3 is denoted as 4-branch AR. As listed in Tab. 2, we form the SFS anchor design using 4-branch AR with the BS of 4. Our method extends SSH in some aspects. Similar to SSH, SFA has the equivalent network architecture in branch M1, M2, and M3. When the BS of anchors is set to 4, the AR of 1 and 2 are deployed in detection module M0 so that the smallest anchor in our method starts with 4. Thus, plenty of small anchors are densely tiled on the image. However, these small anchors inevitably lead to a sharp increase in the number of negative anchors on the background. Thanks to OHEM [29], SFA can balance the positive and negative anchors with a ratio of 1:3 in each mini-batch. Mining hard samples in training is critical to strengthen the power of detector.
\subsection{Feature Maps Fusion Strategy}
Small faces are difficult to be detected not only because of their small scale. Atypical pose, heavy occlusion, extreme illumination, low resolution and other variations in unconstrained scenarios always make CNNs based feature extraction hard to obtain sufficient and complete features for detecting small faces. Therefore, most of small faces become hard faces.
\par To further improve the ability of detecting hard faces, we proposed the Feature Maps Fusion (FMF) strategy. FMF strategy is applied in SFA by partially combining high-level features into low-level features. We fuse the feature maps of neighboring branch and apply the features coming from larger scale to auxiliary detect small faces according to a bold guess that faces with neighboring scale have similar features. We use the FMF strategy in branch M0 and M1 as seen in Fig. 1(b), which receive the early extracted feature maps from Conv3\_3 and Conv4\_3. FMF modules are offered to combining high-level features into low-level features. Fig. 2 shows the architecture of FMF strategy. More precisely, feature maps \textit{M$_{i+1}$} are upsampled and summed up with feature maps \textit{M$_{i}$} where \textit{i} $\in$ \{0, 1\}, followed by a 3$\times$3 convolutional layer. We used bilinear upsampling in the fusion process.
\par By using FMF strategy, SFA is robust to different kinds of variations for small faces to some extent, including occlusion, illumination, low resolution, blur, etc. Benefit from the feature maps coming from neighboring branch with larger scale, SFA can also detect small faces well even though the feature maps in current branch are insufficient and incomplete due to different kinds of variations. From the results of ablation study in Section 4.3.3, we can see that FMF strategy significantly improves the detection performance on the hard set of WIDER FACE [30] dataset which includes a lot of small faces.
\par In fact, medium scale faces contain sufficient and complete features by feature extraction based on CNNs. Therefore, there is no need to fuse the feature maps between medium and large faces. Ablation study in Section 4.3.3 also shows that FMF strategy is not fit for medium faces.

\subsection{Multi-scale Training and Testing}
Instead of using a fixed scale in both training and testing phase, we perform Multi-scale Training (MS-Training) and Multi-scale testing (MS-Testing) strategy to learn more features across a wide range of scales, which makes our model more robust towards different scales and significantly improves the detection performance.
\par In the training phase, we first resize the shortest side of the input image \textit{I} up to \textit{S$_{i}$} (\textit{S$_{i}$} $\in$ \textit{S}) while keeping the largest side below Max Size (1600 in our method). Then, we scale the image according to \textit{S} in MS-Training. For example, when the scale \textit{S} of MS-Training is set to 500, 800, 1200, and 1600, denoted as 4-scale, the input image \textit{I} is first resized to 1200$\times$1600, then we scale the resized image with the size of 500, 800, 1200, and 1600 in the pyramid. In the testing phase, MS-Testing is performed accordingly. We build an image pyramid with a wider range of scales for each test image. Limited to the capacity of GPU memory, the scales of 500, 600, 700, 800, 900, 1000, 1100, 1200, and 1600, denoted as wide-scale, are applied in multi-scale testing phase. Each scale in the pyramid is independently tested. The detection results from various scales are eventually merged together as the final result \textit{D$_{f}$} of the input image \textit{I} as shown in Fig. 1(a).
\par MS-Training makes parameters of four detection modules (detection module M0, M1, M2, and M3) in SFA robust to detect faces with various scales as illustrated in Tab. 2. Different detection modules focus on its own attention scale of faces. As MS-Testing is used in the testing phase, each face of input image \textit{I} will be rescaled accordingly. These rescaled faces may be detected by SFA from different detection modules whose attention scales match with the size of rescaled faces. If at least one rescaled face is found by certain detection module, the original face in input image \textit{I} is successfully detected. 
\par Benefit from MS-Training and MS-Testing, SFA enlarges small faces and easily detect them in medium and large anchors. Fig. 3 shows an example of using MS-Testing. The table in Fig. 3 lists different detection result \textit{D$_{i}$} of rescaled \textit{P$_{i}$}. These detection results are merged to generate the left image as its final detection result \textit{D$_{f}$}. We denote \textit{F = \{F$_{1}$, F$_{2}$, F$_{3}$, F$_{4}$, F$_{5}$, F$_{6}$\}} as the face collection of final detection result \textit{D$_{f}$}. Face \textit{F$_{3}$} and \textit{F$_{4}$} are small faces while they can be detected from rescaled image \textit{P$_{4}$} by using detection module M2 whose anchors attention faces with medium scale. At the same time, SFA shrinks large faces and better detects them in small and medium anchors as well as rescales and finds medium faces with the help of small and large anchors to some extent. As seen in Fig. 3, face \textit{F$_{5}$} is medium face but they can be detected from rescaled image \textit{P$_{4}$} by using detection module M3 whose anchors attention faces with large scale.
\par Though SFA is scale-invariant by design, image pyramid can also improve the performance in both training and testing phase. Ablation study in Section 4.3.3 shows that MS-Training can enhance the detection performance on all subsets, especially on the hard set. Surprisingly, the runtime of SFA will not increase if we adopt MS-Training. Hence, we denoted it as our real-time SFA face detector which adopts MS-Training strategy only. Besides, MS-Testing can improve the detection performance on all subsets by a large margin. Therefore, we deploy both MS-Training and MS-Testing strategy in our final SFA face detector model.
\section{Experiments}
In this section, we firstly analyze the effectiveness of our proposed strategies with comprehensive ablative experiments. Then, we evaluate the final optimal model and achieve state-of-the-art results on common face detection benchmarks. The inference time is finally presented.
\subsection{Experimental Setup}
The parameters of SFA networks are initialized from a pre-trained ImageNet classification model. Our method fine-tunes the resulting model using stochastic gradient descent (SGD) with 0.9 momentum and 0.0005 weight decay. The maximum number of iterations is 54k and stepsize is 18k. The learning rate is firstly set to 0.004 and decreases by a factor of 0.1. Anchors with IoU greater than 0.45 are assigned to positive class and anchors which have IoU less than 0.35 with all ground-truth faces are assigned to the negative class while the rest are ignored. For anchor generation, we use AR of \{1, 2\} in M0, \{4, 8\} in M1, \{16, 32\} in M2, and \{64, 128\} in M3 with a BS of 4. All anchors have aspect ratio of one. Each training image uses horizontal flipping with probability of 0.5 as our data augmentation strategy. We employ the multi-task loss as our objective function. Besides, online negative and positive mining (OHEM) [29] is applied to balance the positive and negative training examples with a ratio of 1:3. During training, 256 detections per module are selected for each image. During inference, each module outputs 1000 best scoring anchors as detections and NMS with a threshold of 0.3 is performed on the outputs of all modules together. Our method is implemented in Caffe [32] and all the experiments are trained on 2 NVIDIA GeForce GTX 1080Ti GPUs in parallel. The code will be available to the research community.
\subsection{Datasets}
WIDER FACE dataset [30]: This dataset contains 32,203 images with 393,703 labeled faces with a high degree of variability in scale, pose and occlusion. It is organized based on 61 event classes, which have much more diversities and are closer to the real-world scenarios. The images in this dataset are split into training (40\% and 12880 images), validation (10\% and 3226 images), and testing (50\% and 16097 images) set. Thus, 158989 labeled faces are in the training set, while 39496 in the validation set and the rest in the testing set. Faces in this dataset are classified into Easy, Medium, and Hard subsets according to the difficulties of detection. The hard subset includes a lot of small faces. Mean average precision (mAP) score is used as the evaluation metric. Plotting scripts for generating the precision-recall (PR) curves are provided to evaluate the performance on the validation set online. While evaluating on the testing set, the results are needed to be sent to the dataset server for receiving the PR curves. We train all models on the training set of the WIDER FACE dataset and evaluate on its validation and test sets. Ablation studies are also performed on the validation set.
\par FDDB dataset [31]: It contains the annotations for 5171 faces in a set of 2845 images taken from news articles on Yahoo websites. Most of the images in the FDDB dataset have less than 3 faces that are clear or slightly occluded. The faces generally have large sizes and high resolutions compared to WIDER FACE. Instead of rectangle bounding boxes, faces in FDDB are represented by bounding ellipses. We use the same model of Experiment XIII presented in Section 4.3 which trained on WIDER FACE training set to perform the evaluation on the FDDB dataset.

\subsection{Ablation Study}
We conduct ablation experiments to examine how each of these proposed strategies affects the final performance. The detailed experimental results of the ablation studies are listed in Tab. 3.

\subsubsection{Baseline setup}
Our method extends from SSH which is a single stage anchor-based face detector. The framework of SSH consists of 3-branch detection architecture (branch M1, M2, and M3), feature maps fusion module M1, and 3-branch AR (\{1, 2\} in M1, \{4, 8\} in M2, and \{16, 32\} in M3) with a BS of 16 in three detection modules. In order to obtain a simple 3-branch detection architecture, we remove FMF module M1 from SSH and use the rest of it as our baseline detector as listed in Tab. 3.
\subsubsection{Ablation setting}
First of all, to better understand the impact of 4-branch detection architecture, we add a new branch M0 on the baseline to form 4-branch detection architecture and denote it as Experiment I. To be fair, branch M0 use the same AR in detection module as branch M1 does (e.g., \{1, 2\} in both M0 and M1). All other factors are the same.
\par Second, we evaluate the effect of SFS anchor design. For anchor generation, we use 4-branch AR (e.g., \{1, 2\} in M0, \{4, 8\} in M1, \{16, 32\} in M2, and \{64, 128\} in M3) but with different BS of 16, 8, and 4 in Experiment II, III, and IV separately. All of these experiments are based on 4-branch detection architecture like Experiment I. Other parameters remain the same.
\par Third, by further examining the impact of FMF strategy, we add the FMF module M0, M1, M0M1, and M0M1M2 in experiment V, VI, VII, and VIII respectively. All of these experiments are based on the detection architecture of Experiment IV.
\par Fourth, we evaluate the influence of MS-Training and MS-Testing. At First, we exploit MS-Training in Experiment IX, which is based on Experiment VII. Similar to SSH, 4-scale (e.g., 500, 800, 1200, and 1600) is used in MS-Training. Next, we apply 4-scale MS-Testing in Experiment X and XI based on Baseline and Experiment VII. Then, both MS-Training and MS-Testing are deployed in Experiment XII, also based on experiment VII, with the same 4-scale mentioned above. Finally, compared to Experiment XII, a wider range of scales are used in Experiment XIII for MS-Testing. Limited to the capacity of GPU memory, wide-scale (e.g., 500, 600, 700, 800, 900, 1000, 1100, 1200, and 1600) is selected.
\subsubsection{Ablation results}
\textbf{4-branch detection architecture is better.} Compared to the 3-branch baseline in Tab. 3, 4-branch detection architecture in Experiment I slightly improves the detection performance on the hard set (rising by 0.3\%) and even on the easy and medium set. The result of Baseline and Experiment I show that 4-branch detection architecture is better for improving the detection performance, especially on the hard set. Therefore, the following ablation studies will adopt the 4-branch detection architecture.
\par \textbf{Small Faces Sensitive anchor design is crucial for detecting small faces.} The comparison among the result of Experiment II, III, and IV in Tab. 3 indicates that SFS anchor design is crucial. With the decrease of BS in anchor design, the detection performance gradually improves on the easy, medium, and hard set. Specifically, when BS is set to 4, the lowest anchor scale in Experiment IV is 4 which differ from 16 in Baseline. And the anchor ratios of 1, 2, 4, and 8 with the BS of 4 are applied for detection module M0 and M1 to jointly deal with small faces. Besides, compared to Experiment I, the result in Experiment IV is slightly lower on all validation sets. Though the same AR of \{1, 2\} in detection module M0, BS is 16 in Experiment I but 4 in Experiment IV, leading to different anchor sizes. Smaller anchors make it possible to find some more small faces at the cost of the rising of mistake rate. Thanks to the FMF strategy mentioned in Section 3.4 above, we can decrease the mistake rate in the following ablation studies. More experiments indicate that SFS anchor design in Experiment IV provides more potential room to improve the detection performance. In order to achieve an elegant anchor design, we will adopt SFS anchor design like Experiment IV in the following ablation studies.
\par \textbf{Feature Maps Fusion strategy is promising for detecting hard faces.} From the results of Experiment VII in Tab. 3, we can see that the detection performance has a great improvement, especially on hard set (about 0.9\% compared to Experiment IV), by using FMF module M0 and M1 simultaneously. Surprisingly, the detector with FMF is robust to different kinds of variations to some extent, including occlusion, illumination, blur, etc. However, when FMF module M0, M1, and M2 is used in Experiment VIII, the detection performance sharply drops on the hard set. Compared to Experiment IV without feature maps fusion, the detection performance in Experiment VIII is worse on the hard set (about 0.9\%). So, we will use FMF module M0 and M1 in the following ablation studies.
\par \textbf{Multi-scale Training and Testing can significantly improve the detecting performance.} MS-Training is applied in Experiment IX with the 4-scale (e.g., 500, 800, 1200, 1600) based on Experiment VII. The result of Experiment IX shows that MS-Training is helpful for enhancing the detection performance, especially on the hard set. We denoted it as our real-time SFA face detector which adopts MS-Training strategy. Benefit from MS-Testing, the detection performance of Experiment X and XI have a great improvement on all validation sets compared to Baseline and Experiment VII. Later, Experiment XII adopts both MS-Training and MS-Testing with the same 4-scale simultaneously and further improves the detection performance. Finally, wide-scale (e.g., 500, 600, 700, 800, 900, 1000, 1100, 1200, 1600) is used in Experiment XIII for MS-Testing. Compared to Experiment VII, the result of Experiment XIII increases 2.2\%, 2.1\%, and 3.6\% on the easy, medium, and hard set separately, which demonstrates that MS-Training and MS-Testing can significantly improve the detecting performance. 
\par Combining all the above strategies achieves the best detection performance (as shown in Experiment XIII) and denotes it as our final SFA detector model.

\subsection{Evaluation on Benchmark}
We evaluate our proposed method against state-of-the-art methods on two public face detection benchmarks (i.e. WIDER FACE [30] and FDDB [31]).
\subsubsection{WIDER FACE dataset}
Our method is trained on the training set of the WIDER FACE dataset and evaluate on its validation and testing set against the recently published state-of-the-art face detection methods including Zhu et al. [27], S$^{3}$FD [21], SSH [20], HR [26], MSCNN [33], CMS-RCNN [24], Multitask Cascade CNN [18], LDCF+ [34] and Multiscale Cascade CNN [30]. The precision-recall curves and mAP values on WIDER FACE validation and testing sets are presented in Fig. 4. As can be seen, the proposed SFA approach based on a headless VGG-16 consistently achieves the impressive performance across all the three subsets, especially on the hard subset which mainly contains small faces. It achieves the promising average precision in all level faces, i.e. 0.949 (Easy), 0.936 (Medium), and 0.866 (Hard) for validation set, and 0.941 (Easy), 0.930 (Medium), and 0.862 (Hard) for testing set. In particular, SFA outperforms most of these methods by a large margin except method proposed by Zhu et al. [27]. When evaluated on the WIDER FACE validation set, our method achieves better performance against most of prior methods while has the same performance as method of Zhu et al. [27] on the easy subset. Besides, our SFA is consistently better than most of other state-of-the-art methods and only poorer than method of Zhu et al. [27] on all three subsets when evaluated on the WIDER FACE testing set. One possible reason for the relatively poor performance of SFA against method of Zhu et al. [27] is that SFA uses VGG-16 as its backbone network while method proposed by Zhu et al. [27] uses ResNet-101 [8]. The result in Fig. 4 not only demonstrates the effectiveness of the proposed method but also strongly shows the superiority of the proposed model in detecting small and hard faces.

\subsubsection{FDDB dataset}
In these datasets, we resize the shortest side of the input images to 400 pixels while keeping the larger side less than 800 pixels, leading to an inference speed of more than 20 FPS. And we directly use our final SFA detector model in Experiment XIII and compare SFA against the recently published state-of-the-art methods including FD-CNN [35], ICC-CNN [36], RSA [37], S$^{3}$FD [21], FaceBoxes [22], HR [26], HR-ER [26], DeepIR [38], LDCF+ [34], UnitBox [39], Conv3D [40], Faster RCNN [41] and MTCNN [18] on FDDB dataset. For a more fair comparison, the predicted bounding boxes are converted to bounding ellipses. Fig. 5(a) and Fig. 5(b) show the discrete ROC curves and continuous ROC curves of these methods on the FDDB dataset respectively. The proposed SFA approach consistently achieves the impressive performance in terms of both the discrete ROC curves and continuous ROC curves. For the discrete ROC score, our face detector outperforms most of these methods by a large margin except S$^{3}$FD detector, as shown in Fig. 5(a). With the more restrictive continuous scores, our SFA is better than most of other state-of-the-art methods while poorer than S$^{3}$FD, DeepIR, and RSA, as shown in Fig. 5(b). It should be noted that the performance of S$^{3}$FD is achieved after manually adding 238 unlabeled faces on the test set. And we use the same annotation on FDDB dataset offered by S$^{3}$FD. Furthermore, except for the WIDER FACE training set, RSA uses 171K images collected from the Internet and DeepIR uses the WIDER FACE validation set in the training phase, while SFA only uses the WIDER FACE training set. These results demonstrate the effectiveness and good generalization capability of SFA to detect unconstrained faces.

\subsection{Inference Time}
\par In this section, we report the inference time of our proposed SFA face detector on the WIDER FACE validation set. Benefit from the single stage of SFA, our inference time is independent of the number of faces in an image. It can detect faces from images with arbitrary size. Specifically, the inference time of our proposed method is determined by three aspects as follows: (1) Max scale \textit{S$_{max}$} in MS-Testing; (2) Max Size; (3) The number of scales \textit{N$_{s}$} in MS-testing. A feature map is a tensor of size \textit{C}$\times$\textit{W}$\times$\textit{H}, where \textit{C} is the number of channels, \textit{W} and \textit{H} are the width and the height respectively. When \textit{S$_{max}$} and Max Size are set, the size of largest feature map in SFA is determined at the same time. More precisely, \textit{W} is set to \textit{S$_{max}$} and \textit{H} is set to Max Size.

\par The speed is measured by using NVIDIA GeForce GTX 1080Ti GPU and cuDNN v5.1 with Intel Core i7-6850k CPU@3.60GHz. Tab. 4 shows the inference time with respect to the size of largest feature map (W$\times$H) and the number of scales N$_{s}$ in MS-Testing. It can be observed that our face detector runs at 20 FPS with 480$\times$640 largest feature map and 10 FPS with 720$\times$1280 largest feature map, as the row 2 and 4 listed in Tab. 4. For the 1200$\times$1600 largest feature map with batch size 1 using a single GPU, our real-time SFA face detector can run at 5 FPS as well as maintain high performance, as the row 5 listed in Tab. 4. When multiple scales are deployed in MS-Testing, the inference time would be dominated by the size of largest feature map and the number of scales \textit{N$_{s}$}. Our final SFA detector as the last row of Tab. 4 described can take 1.4s to detect faces from an image with 1600$\times$1600 largest feature map. From Experiment XII in Tab. 3 and the last second row in Tab. 4, we can see that when using MS-Training and MS-Testing with the same 4-scale, our method can take 0.75s to detect faces from an image with 1600$\times$1600 largest feature map. It achieves slightly lower detection performance against our final SFA detector but reduces half of inference time. Therefore, we denote SFA using MS-Training and MS-Testing with 4-scale as our real-world SFA face detector. In fact, most of the forward time is spent on the VGG-16 network, which limits the inference speed. To speed up the inference time, we will adopt ResNet-101 as our backbone network to accelerate face detector in our future work.
	
\subsection{Qualitative Results}
Fig. 6 shows some examples of the face detection results using the proposed SFA on the WIDER FACE validation dataset. Fig. 6(a) lists some difficult cases. Our method is able to detect faces with different scales, especially for small faces (see the first row in Fig. 6(a)). Besides, SFA can also achieve satisfied detection results on hard faces caused by atypical pose, heavy occlusion, exaggerated expression, make up, extreme illumination and blur (see the last two rows in Fig. 6(a)). Fig. 6(b) lists some selected false positives. In fact, most of the false positives in SFA are actually human faces caused by missing labels (see the first two rows in Fig. 6(b)). For other false positives, we find errors made by our model are rather reasonable. They all have the pattern of human face and fool our model to treat it as a face (see the last row in Fig. 6(b)).

\par Fig. 7 shows some examples of the face detection results generated by SFA on the FDDB dataset. Fig. 7(a) lists some difficult cases including faces with different scale, atypical pose, heavy occlusion, exaggerated expression, and blur. Benefit from excellent performance of SFA in detecting small faces and hard faces, we can find a lot of faces from human perspective but lack of labels on the FDDB dataset, as seen in Fig. 7(b). Our method is able to find extra faces with small scale which are not labeled (see the first row in Fig. 7(b)). Besides, some faces with atypical pose can also be detected (see the second row in Fig. 7(b)). The detection results of faces with heavy occlusion, blur, and wrong label are shown in the last row of Fig. 7(b).

\section{Conclusion}
In this paper, we propose a novel face detection architecture focus on small faces and present a new face detector to deal with the open problem of anchor-based detection methods whose performance drops sharply as the faces becoming smaller. Multiple strategies are deployed in SFA for the sake of better detecting small faces, such as 4-branch detection architecture, small faces sensitive anchors design, feature maps fusion strategy, and multi-scale training and testing strategy. These strategies make SFA rapid, efficient, and robust to detect faces in unconstrained settings, especially for small faces. Extensive experiments demonstrate that our method outperforms most of the recently published face detectors and achieves the state-of-the-art performance on challenging face detection benchmarks like WIDER FACE and FDDB datasets with competitive inference speed. In our future work, we will adopt ResNet-101 as our backbone network to extract more robust and intrinsic feature of faces as well as accelerate face detector.
\section*{Acknowledgments}
This work was supported by National Science $\&$ Technology Pillar Program of China (Grant No. 2012BAH48F02), National Natural Science Foundation of China (Grant No. 61801190), Natural Science Foundation of Jilin Province (Grant No. 20180101055JC), Outstanding Young Talent Foundation of Jilin Province (Grant No. 20180520029JH) and China Postdoctoral Science Foundation (Grant No. 2017M611323).
\section*{References}
\label{}







\end{document}